\documentclass[conference]{IEEEtran}
\IEEEoverridecommandlockouts
\usepackage{array}
\usepackage[caption=false,font=footnotesize,labelfont=sf,textfont=sf]{subfig}
\usepackage{amsthm}
\usepackage{color}
\usepackage{booktabs}
\usepackage[ruled,linesnumbered]{algorithm2e}
\usepackage{float}
\usepackage{multirow}
\usepackage{multicol}
\usepackage{soul}
\usepackage{bm}
\usepackage{marvosym}
\usepackage{url}
\usepackage{graphicx}
\usepackage{subfloat}
\usepackage{caption}
\usepackage{float}
\usepackage{flushend}
\usepackage{enumitem}
\usepackage{hyperref}

\newcommand\degree{^\circ}
\newcommand{\myfw}{AUTO}
\newcommand{\tabincell}[2]{\begin{tabular}{@{}#1@{}}#2\end{tabular}}
\usepackage{cite}
\usepackage{amsmath,amssymb,amsfonts}
\usepackage{graphicx}
\usepackage{textcomp}
\usepackage{xcolor}

\def\BibTeX{{\rm B\kern-.05em{\sc i\kern-.025em b}\kern-.08em
    T\kern-.1667em\lower.7ex\hbox{E}\kern-.125emX}}
\begin{document}

\title{Parameterized Decision-making with Multi-modal Perception for Autonomous Driving}

\author{Yuyang~Xia\textsuperscript{1}, Shuncheng~Liu\textsuperscript{1}, Quanlin~Yu\textsuperscript{2},
Liwei~Deng\textsuperscript{1}, You~Zhang\textsuperscript{3}, Han~Su\textsuperscript{2,1}\textsuperscript{\Letter}\thanks{\textsuperscript{\Letter}Corresponding authors: Han Su and Kai Zheng.}, Kai~Zheng\textsuperscript{1}\textsuperscript{\Letter}\\
\textsuperscript{1}University of Electronic Science and Technology of China, China \quad \\
\textsuperscript{2}Yangtze Delta Region Institute(Quzhou), University of Electronic Science and Technology of China \\
\textsuperscript{3}University of Michigan, USA\\
\{xiayuyang,liushuncheng,quanlin.yu,deng\_liwei\}@std.uestc.edu.cn, \\
youzh@umich.edu, \{hansu,zhengkai\}@uestc.edu.cn
}


\maketitle

\begin{abstract}
Autonomous driving is an emerging technology that has advanced rapidly over the last decade. Modern transportation is expected to benefit greatly from a wise decision-making framework of autonomous vehicles, including the improvement of mobility and the minimization of risks and travel time. However, existing methods either ignore the complexity of environments only fitting straight roads, or ignore the impact on surrounding vehicles during optimization phases, leading to weak environmental adaptability and incomplete optimization objectives. To address these limitations, we propose a p\underline{A}rameterized decision-making framework with m\underline{U}lti-modal percep\underline{T}i\underline{O}n based on deep reinforcement learning, called \myfw. We conduct a comprehensive perception to capture the state features of various traffic participants around the autonomous vehicle, based on which we design a graph-based model to learn a state representation of the multi-modal semantic features. To distinguish between lane-following and lane-changing, we decompose an action of the autonomous vehicle into a parameterized action structure that first decides whether to change lanes and then computes an exact action to execute. A hybrid reward function takes into account aspects of safety, traffic efficiency, passenger comfort, and impact to guide the framework to generate optimal actions. In addition, we design a regularization term and a multi-worker paradigm to enhance the training. Extensive experiments offer evidence that \myfw\ can advance state-of-the-art in terms of both macroscopic and microscopic effectiveness.
\end{abstract}


\section{Introduction\label{introduction}}
With the rapid growth in urbanization and vehicle ownership, most major cities around the world suffer from traffic congestion, resulting in serious traveling inefficiency, fuel waste, and air pollution~\cite{schrank2021urban}. Typically, road environments or drivers are to blame for traffic congestion~\cite{won2016toward}. Environment variables like road construction and a reduction in the number of lanes (bottleneck) can give rise to traffic congestion. Additionally, a driver’s poor driving behaviors (e.g., hard braking and abrupt lane-changing) may result in traffic congestion or even accidents.
However, the latter is more frequent due to some limitations of human drivers (limited field of view and long reaction time) and heterogeneity (different driving habits among drivers)~\cite{qu2020jointly}.
It is highly challenging for human drivers to make optimal decisions all the time.

In the last decade, autonomous driving has gained broad attention from the public, which aims to assist or even replace a human driver with a robot that constantly receives environmental information via various sensor technologies (as compared to human eyes) and consequently determines vehicle behaviors with proper algorithms (as compared to human brains)~\cite{merriman2021challenges}. 
Considering a mechanism that perceives the surrounding traffic environment and makes decisions, it fits well within the realm of reinforcement learning~\cite{kiran2021deep}. There have been many works utilizing reinforcement learning-based methods~\cite{leurent2019social,aradi2020survey,nageshrao2019autonomous} to accomplish autonomous driving and outperform traditional rule-based methods~\cite{xiao2017realistic,erdmann2015sumo,milanes2014modeling} and deep learning-based methods~\cite{prakash2021multi,chen2019deep,tedjopurnomo2020survey}.
However, the existing works share some common limitations, including \emph{weak environmental adaptability} and \emph{incomplete optimization objectives}.
For the first limitation, the aforementioned algorithms are implemented in simple traffic scenarios (e.g., highways). They assume that the environment is constant and vehicle information can be accurately acquired in advance, leading to low applicability and high uncertainty in complex traffics. Recently, some methods utilize cameras and/or LiDARs to perceive complex environmental information to adapt to autonomous driving on urban roads~\cite{fu2022decision,9835644,perez2022deep}.
However, they over-simplify the environments only using images and/or point cloud data that ignore road maps and varying states of the autonomous vehicle, e.g., orientation and offset from lanes.
Although multi-sensor fusion is the trend of autonomous driving, few studies have fully exploited multi-modal information from sensors and road networks that can improve adaptability.
For the second limitation, existing approaches~\cite{fu2022decision,zhu2020safe,xu2021patrol} mainly optimize the driving safety, traffic efficiency, and passenger comfort of autonomous vehicles, leaving the impact on other conventional vehicles largely uninvestigated.
Generally, hard braking and forced lane-changing behaviors of a vehicle can cause a negative impact on its rear vehicles, which eventually causes traffic delays and even congestion~\cite{talebpour2016influence,stern2018dissipation}.  
The former framework~\cite{9835644} attempts to make coarse-grained decisions with consideration of multiple impact situations. 
Its lane-changing behaviors are discrete rather than continuous steering angle control. Therefore, this method are not practical for real-world driving situations and cannot deal with the impact of fine-grained steering angle control. Further, the discrete manner cannot roll back a lane-changing decision halfway, which may cause safety issues in some corner cases.



To enhance environmental adaptability, we can consider a comprehensive traffic environment including roads, vehicles, traffic signs, and traffic lights. The autonomous vehicle is able to collect information from different data sources including an offline high-definition map (HD map) and multiple onboard sensors~\cite{gomez2022build,feng2020deep,gomez2020train}, i.e., Camera, Light Detection and Ranging (LiDAR), Global Navigation Satellite System (GNSS), and Inertial Measurement Unit (IMU). 
Specifically, HD maps provide lane information (e.g., topology and curvature) and static traffic regulatory elements (e.g., traffic signs) along a specific route, and Cameras capture time-varying traffic light states, and LiDARs obtain the information (e.g., velocity, position, and size) of surrounding vehicles, and GNSS and IMU provide the position and orientation of the autonomous vehicle.
In this context, the autonomous vehicle gets multi-modal inputs from HD maps and sensors, and a decision-maker outputs velocity and steering angle at each time step. To optimize decision-making, we can enable the autonomous vehicle to complete a specific route with driving safety, traffic efficiency, passenger comfort, and minimal impact on surrounding vehicles following the paradigm of reinforcement learning.
However, the intuitions will face three main challenges:
(1) Environmental information is complex and multi-modal. The aggregation methods in previous works~\cite{9835644,fu2022decision} are time-consuming and fail to exploit road semantic information since they model the multi-modal traffic participants as a complete graph and ignore the distinctions in traffic conditions on different lanes. 
(2) Both lane-following and lane-changing require calculating steering angles. Only generating a steering angle usually causes the autonomous vehicle to deviate far from the lane centerline since it cannot distinguish between lane-changing behaviors and lane-following behaviors on curved multi-lane roads.
(3) It is common to optimize the safety, traffic efficiency, and passenger comfort of the autonomous vehicle~\cite{zhu2020safe,fu2022decision,xu2021patrol}, but minimizing the impact on surrounding vehicles has been less studied. How to design reward functions and measure the impact factor is challenging.


To tackle these challenges, we propose a p\underline{A}rameterized decision-making framework with m\underline{U}lti-modal percep\underline{T}i\underline{O}n based on deep reinforcement learning, called \myfw. To address the first challenge, we propose a state representation model LCA that aims to learn a comprehensive representation of the multi-modal features. It organizes the multi-modal features as multiple agent-centric star graphs to efficiently represent their interaction relationship, and proposes a novel lane-wise fuse paradigm to effectively exploit road semantic information. To address the second challenge, we propose a reinforcement
learning model RBP-DQN to optimize a parameterized action structure, which calculates three different sets of action values (i.e., steering angle and acceleration brake rate) to respectively represent left lane-changing, lane-following, and right lane-changing behaviors and then selects the one with the greatest reward value to execute. This design allows the autonomous vehicle to adjust the steering
angle accurately, thus achieving better driving performance. In addition, due to the complexity of the traffic environment and action space, we design a regularization term to prevent large policy shifts during policy optimization. To address the third challenge, we design a hybrid reward function to guide our framework to learn the state representation and optimize the parameterized action calculation. It incorporates safety, traffic efficiency, passenger comfort, and impact terms. In particular, the impact term is used to penalize the autonomous vehicle when it executes harsh actions that force other vehicles to decelerate, thus reducing the impact of the autonomous vehicle on traffic flow.

To the best of our knowledge, this work provides the first data-driven solution to make decisions in continuous action space that considers multi-modal information inputs and the impact of the autonomous vehicle. Overall, our main contributions can be summarized as follows:

\noindent$\bullet$ We develop a decision-making framework that enables the autonomous vehicle to complete a route in comprehensive traffic with safety, traffic efficiency, passenger comfort, and minimal impact on surrounding vehicles.

\noindent$\bullet$ We propose an efficient graph-based model to exploit valuable features from multi-modal environmental data and design a parameterized action paradigm to calculate fine-grained actions based on coarse-grained decisions.

\noindent$\bullet$ We propose a hybrid reward function to guide the optimization of reinforcement learning and use a regularization term and a multi-worker setting to enhance the training.


\noindent$\bullet$ We conduct extensive experiments to evaluate the superiority of our framework on an open-source simulator from both macroscopic and microscopic metrics.

\section{Problem Definition}

\begin{figure*}[!t]
    \centering
    \setlength{\abovecaptionskip}{0.1cm}
    \includegraphics[width=0.98\textwidth]{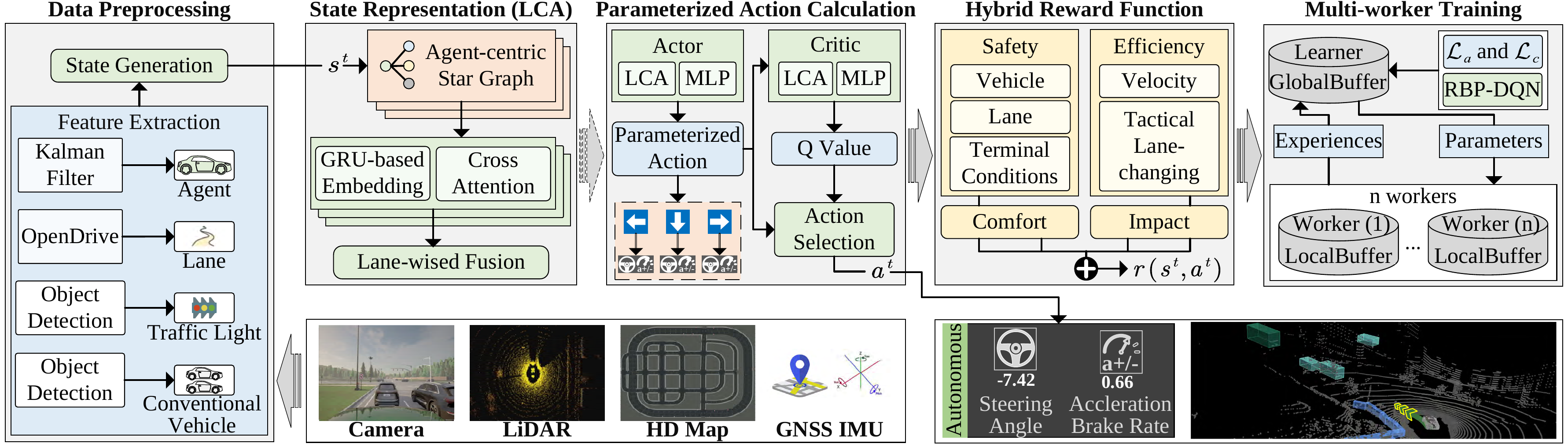}
    \caption{Framework Overview}
    \label{framework}
    \vspace{-0.6cm}
\end{figure*}

\subsection{Preliminary Concepts}
\label{definition}
We consider an interactive traffic environment where there is one autonomous vehicle $A$ and a set of conventional vehicles $\mathbb{C}$ traveling on complex multi-lane roads. The autonomous vehicle is equipped with an HD map and onboard sensors, i.e., Camera, LiDAR, GNSS, and IMU. Benefiting from the development of feature extraction techniques~\cite{gomez2022build,feng2020deep,gomez2020train} for map and sensor data, we can obtain the real-time multi-modal features of lanes, vehicles, and traffic lights along a specific route. Based on these features, the autonomous vehicle makes an action decision (i.e., a steering angle and an acceleration brake rate) at each time step $t$ within a time duration $\mathbb{T}$ of interest. 
Next, we will explain some definitions and notations used in the rest of this paper.

\noindent \underline{\textbf{Route.}}
A route $\mathit{rou}$ is a sequence of roads that the autonomous vehicle needs to follow in succession, which indicates a planned path from an origin to a destination.
It can be denoted as $\footnotesize \mathit{rou}= \left \langle \mathit{rd}_1, \mathit{rd}_2,\dots,\mathit{rd}_g  \right \rangle$, where $g$ is the number of roads in the route.

\noindent \underline{\textbf{Lane.}} 
A lane $l$ is part of a road used to guide drivers and reduce traffic conflicts. Usually, a road has multiple lanes, and they are numbered incrementally from the leftmost side to the rightmost side, i.e., $\footnotesize l_1, l_2,\dots, l_{\mathcal{K}}$, where $l_1$ and $l_\mathcal{K}$ indicate the leftmost lane and rightmost lane, respectively.
Each lane is represented as a sequence of waypoints sampled equidistantly on the centerline at one-meter intervals~\cite{chang2019argoverse}, and the feature vector $f_{wp}$ of each waypoint $\mathit{wp}$ is defined as $\footnotesize f_\mathit{wp}=(\mathit{alt}, k, \theta, \mathit{attr})$ that consists of the altitude, lane ID, orientation, and attribute of the waypoint. The altitude and orientation provide the slope and curvature of the waypoint, and the attribute refers to the static traffic sign information at the waypoint. We consider speed limit signs for now and set $\mathit{attr}$ of each waypoint as the speed limit.

\noindent \underline{\textbf{Vehicle.}}
We represent the features of the autonomous vehicle as $\footnotesize f_{A}=(\mathit{vel}, \theta, \mathit{len}, \mathit{wid}, \mathit{off})$, where $\footnotesize f_{A}.\mathit{vel}$ is the velocity, $\footnotesize f_{A}.\mathit{\theta}$ is the orientation, $\footnotesize f_A.\mathit{len}$ is the length, $\footnotesize f_A.\mathit{wid}$ is the width, and $\footnotesize f_A.\mathit{off}$ denotes the offset from the current lane centerline~\cite{gutierrez2020waypoint}. Then for the conventional vehicles $\mathbb{C}$, we calculate their relative features with respect to the autonomous vehicle $A$. For example, the features of $C_i$ are denoted as $\footnotesize f(C_{i}, A)=(\mathit{dis}, \mathit{vel}, \theta, \mathit{len}, \mathit{wid}, \mathit{off})$, where $f(C_{i}, A).\mathit{dis}$ refers to the relative distance along the lane, $\footnotesize f(C_{i}, A).\mathit{vel}$ and $\footnotesize f(C_{i}, A).\theta$ refer to the relative velocity and orientation, $\footnotesize f(C_i, A).\mathit{len}$ and $\footnotesize f(C_{i}, A).\mathit{wid}$ denote the length and width, and $\footnotesize f(C_i, A).\mathit{off}$ is the offset from the current lane centerline.

\noindent \underline{\textbf{Traffic Light.}} 
A traffic light $\mathit{tl}$ consists of three signals (i.e., red, yellow, and green) that transmit real-time traffic regulation information to drivers, and vehicles are allowed to pass when the light is green. Formally, The features $f_\mathit{tl}$ of a traffic light $\mathit{tl}$ can be represented as $\footnotesize f_\mathit{tl}=(\mathit{status}, \mathit{dis})$, where $\mathit{status}$ denotes the color of $\mathit{tl}$, and $\mathit{dis}$ denotes the relative distance between the traffic light $\mathit{tl}$ and the autonomous vehicle $A$. Specifically, we use one-hot vectors to represent the status of traffic lights. $\left[1, 0, 0\right]$, $\left[0, 1, 0\right]$ and $\left[0, 0, 1\right]$ denote the red light, yellow light, and green light, respectively.

\noindent \underline{\textbf{Time Step.}} 
In order to model the problem more concisely, we treat the continuous time duration $\mathbb{T}$ as a discretized set of time steps, i.e., $\footnotesize \mathbb{T}=\left \langle 1, 2, \dots, t, \mathit{t+1}, \dots \right \rangle$. We denote $\Delta t$ as the time interval between two consecutive time steps, which is set to 0.1 seconds following the previous works~\cite{liu2022multi,zhu2020safe}.

\noindent \underline{\textbf{Steering Angle.}} 
The steering angle $\mathit{deg}$ determines the orientation of a vehicle, which is defined as the angle between the front of the vehicle and the steered wheel direction. 
The angle varies from $-60\degree$ to $60\degree$~\cite{chen2019attention}. The negative values mean the autonomous vehicle is turning left or changing lanes to the left, and vice versa.


\noindent \underline{\textbf{Acceleration Brake Rate.}} 
The acceleration brake rate $\mathit{abr}$ determines the velocity of a vehicle, which is defined as the magnitude of the accelerator and brake~\cite{chen2019attention}. It is in the range of $\left[-1, 1\right]$, where positive values $\left(0,1\right]$ mean accelerating, negative values $\left[-1,0\right)$ mean braking, and 0 means a neutral state with no braking and accelerating.

\noindent \underline{\textbf{Objective.}}
Our objective is that the autonomous vehicle can output an optimal steering angle $\mathit{deg}$ and an acceleration brake rate $\mathit{abr}$ to follow a specific route while achieving safety, driving efficiency, passenger comfort, and minimal impact on the surrounding conventional vehicles.

\subsection{DRL-based Decision-making}
Considering the mechanism that an autonomous vehicle perceives the surrounding traffic environment and makes a decision, it fits well within the realm of Partially Observable Markov Decision Process (POMDP)~\cite{zhu2021deep} and deep reinforcement learning~\cite{sutton2018reinforcement}. Then we present some important definitions under POMDP as follows: -\ \emph{Agent}: the autonomous vehicle $A$. -\ \emph{Environment}: the surrounding environment. -\ \emph{State $s^{t}$}: current and historical multi-modal inputs extracted from the environment, including the features of the agent and its surrounding lane waypoints, vehicles, and traffic lights along the route. -\ \emph{Action $a^{t}$}: a steering angle $\mathit{deg}^t$ and an acceleration brake rate $\mathit{abr}^t$ of the agent. -\ \emph{Reward $r(s^{t}, a^{t})$}: a feedback value after the agent performs $a^t$ at $s^t$, which is the aggregation of multiple aspects: safety, traffic efficiency, passenger comfort, and impact on rear conventional vehicles. As the agent explores the environment, the deep reinforcement learning paradigm aims to exploit valuable features from states and form a policy that can generate actions with large reward values. 



\section{Framework and Methodology\label{Methodology}}

\subsection{Framework Overview}
Figure~\ref{framework} shows the architecture of our framework, which consists of five components: data preprocessing, state representation, parameterized action calculation, hybrid reward function, and multi-worker training.
For the \emph{data preprocessing}, we take the data from HD maps and multiple sensors (i.e., Camera, LiDAR, GNSS, and IMU) as input, based on which we respectively extract the feature vectors of lane waypoints, vehicles, and traffic lights and generate a multi-modal state $s^t$ for the agent.
For the \emph{state representation}, we propose a lane-wised cross attention model (LCA) to learn a latent representation of the state $s^t$. It organizes $s^t$ as multiple agent-centric star graphs and uses a GRU network to embed the node features. Then, it uses a cross attention mechanism to aggregate each star graph and introduces a lane-wise paradigm to fuse these aggregated results as a state representation.
For the \emph{parameterized action calculation}, we first integrate LCA into an actor network and a critic network, respectively. Then, we compute an action $a^t$ using a parameterized action structure that first decides whether to perform a lane-changing decision (high level) and then compute an exact action to execute (low level).
For the \emph{hybrid reward function}, we calculate a reward value $r(s^t, a^t)$ for the action $a^t$ and state $s^t$, which serves as a signal to guide the agent to learn an optimal action policy. 
For the \emph{multi-worker training}, we use a regularization technique to improve the convergence performance of our reinforcement learning model RBP-DQN and speed up the training speed using distributed computation following the previous studies~\cite{horgan2018distributed}. 

\begin{figure}[!t]
    \centering
    \setlength{\abovecaptionskip}{0.1cm}
    \includegraphics[width=0.48\textwidth]{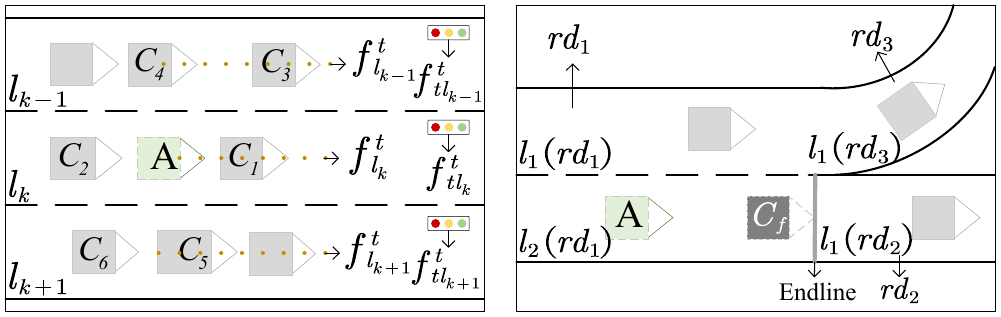}
    \caption{Example of State Generation}
    \label{state generation}
    \vspace{-0.6cm}
\end{figure}

\subsection{Data Preprocessing}
This component extracts multi-modal features (i.e., lane waypoints, vehicles, and traffic lights) from the raw data of HD maps and sensors. Then based on these features and a specific route, we generate a fixed-dimensional state $s^t$ at time step $t$ for the agent.

\noindent \underline{\textbf{Feature Extraction.}}
We conduct the experiments in a widely-used simulator Carla that mimics real-world traffic scenarios. It provides high-fidelity sensor models that can replicate the characteristics (e.g., field of view and resolution) of real-world sensors and generate synthetic data based on the virtual environment and the positions of objects. To obtain the surrounding environment features, we first take advantage of the extended Kalman filter~\cite{ribeiro2004kalman} to process the data from GNSS and IMU to get the position and orientation of the agent, based on which we can locate the agent and the route in an OpenDrive-based HD map~\cite{diaz2022hd} and acquire the surrounding stationary lane features. Then with the aid of object detection technology~\cite{lari2011adaptive,wang2023yolov7}, we extract the features of surrounding conventional vehicles from the point cloud data by LiDAR and extract the features of traffic lights from the image data by Camera. In addition, the detection distances of Camera and LiDAR always have an impact on the decision-making performance of autonomous driving. To test the impact, we conduct a hyper-parameter sensitivity analysis in Section~\ref{Hyper-Parameter Sensitivity}.


\noindent \underline{\textbf{State Generation.}}
\label{state generation text}
Based on the feature extraction, at time step $t$, we can collect the feature vector $f_A^t$ of the agent and $f_{k}^t$, $f_{k-1}^t$, $f_{k+1}^t$ that respectively consists of the multi-modal features on the current lane $l_k$, the left lane $l_{k-1}$, and the right lane $l_{k+1}$. 
As shown in the left side of Figure~\ref{state generation}, we select six conventional vehicles on three lanes, which cover six key areas centered on the agent, i.e., 1) front ($C_1$), 2) rear ($C_2$), 3) front left ($C_3$), 4) rear left ($C_4$), 5) front right ($C_5$), 6) rear right ($C_6$) areas~\cite{liu2021lane,fu2021trajectory}. Then taking $f_k^t$ as an example, $f_k^t$ can be represented as $f_{k}^t=(f_{l_k}^t, f_{v_k}^t, f_{\mathit{tl_k}}^t)$. Specifically, $f_{l_k}^t$ is the concatenation of the feature vectors of waypoints on the lane $l_k$, i.e., $f_{l_k}^t=f_{wp_{k,1}}^t\oplus f_{wp_{k,2}}^t \oplus \dots \oplus f_{wp_{k,m}}^t$,
where $m$ is the number of waypoints. $f_{v_k}^t$ is the concatenation of the feature vectors of vehicle $C_1$ and $C_2$ on $l_k$, i.e., $f_{v_k}^t=f^t(C_1,A)\oplus f^t(C_2,A)$. $f^t_{\mathit{tl_k}}$ refers to the feature vector of traffic light on $l_k$.

In summary, the state $s^t$ consists of $\small F^t_A$ and $\small X^t_{k-1}=[F^t_{l_{k-1}},F^t_{v_{k-1}},F^t_{\mathit{tl}_{k-1}}]$, $\small X^t_{k}=[F^t_{l_k},F^t_{v_k},F^t_{\mathit{tl}_k}]$, and $\small X^t_{k+1}=[F^t_{l_{k+1}},F^t_{v_{k+1}},F^t_{\mathit{tl}_{k+1}}]$. Note that $F$ means that it contains historical feature vectors in the past $z$ time steps rather than just the current time step, e.g., $F^t_{A}=[f_{A}^{t-z+1}, f_{A}^{t-z+2},\dots, f_{A}^t]$.


\subsection{State Representation}
\label{LCA text}
Existing studies on the representation learning of traffic context can be divided into two categories: raster-based methods~\cite{cui2019multimodal,zeng2021modeling} and vector-based methods~\cite{gao2020vectornet,varadarajan2022multipath++}. The raster-based methods rasterize the traffic scene of the agent into a grided image and apply a series of standard convolution layers to process it. However, these approaches require significantly more computation to train and the rasterization process inevitably results in information loss~\cite{liang2020learning}. Recently, the vector-based method as a more succinct representation has developed rapidly in many traffic tasks~\cite{gao2020vectornet,varadarajan2022multipath++}, which uses feature vectors to represent different traffic participants (i.e., lane waypoints, conventional vehicles, and traffic lights) and apply the graph neural network to capture the complex interaction relationship of them. 

Therefore, there are many decision-making works~\cite{fu2022decision,zhu2020safe,xu2021patrol,9835644} increasingly applying vector-based methods to learn a representation vector of the traffic scene. But in this paper, these works usually lead to unsatisfactory results due to the following reasons: (1) Their aggregation manners are redundant since they model the autonomous vehicle and other traffic participants as a complete graph. Although this design can fully exploit the interaction relationship in the traffic, some interactions are irrelevant to the decision-making of our autonomous vehicle, which can introduce additional calculations and degrade the driving performance. (2) They fail to exploit road semantic information in the traffic since they only focus on the aggregation of different traffic participants while ignoring the distinctions in traffic conditions on different lanes. Evidently, an effective road semantic representation can aid the autonomous vehicle to improve its driving decisions, especially lane-changing decisions. (3) Their robustness is poor. In practice, the features collected by the autonomous vehicle are inevitably subject to noises, e.g., the traffic light detection in our experiments has an error rate of up to 18\% compared to the ground truth. Solely relying on the current state features inherently leaves decision-making vulnerable to observation noises. In contrast, incorporating historical features into states empowers the model to comprehend changes in the environment and thus improve the robustness to noises~\cite{moradpopgym}.  

\begin{figure}[!t]
    \centering
    \setlength{\abovecaptionskip}{0.1cm}
    \includegraphics[width=0.46\textwidth]{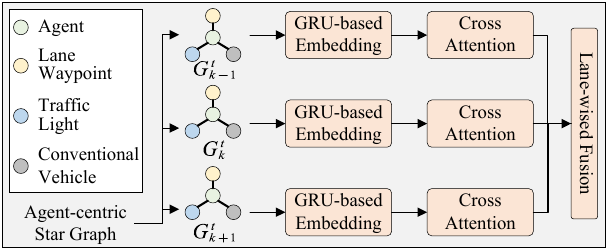}
    \caption{Network Structure of LCA}
    \label{attn}
    \vspace{-0.6cm}
\end{figure}

In this work, we propose a lane-wised cross attention model, called LCA, to learn a representation of the state $s^t$ as shown in Figure~\ref{attn}. To efficiently exploit the interaction relationship, we organize the state $s^t$ into three agent-centric star graphs (i.e., $G^t_{k-1}$, $G^t_k$, and $G^t_{k+1}$) to enable the autonomous vehicle to focus primarily on the influence of different traffic participants on its future decisions. Then, we apply a GRU-based embedding network to embed the historical state features in each node and apply a cross attention layer to aggregate each graph. This step has embedded the road's waypoint features (e.g., position, orientation, and traffic sign features). Finally, we fuse them lane by lane to augment road semantic information.




\noindent \underline{\textbf{Network Structure.}} In this section, we will detail the workflow of LCA as follow:

(1) GRU-based Embedding. Taking $G^t_k$ as an example, $G^t_k$ is an undirected graph $G^t_k=(\mathbb{V}, \mathbb{E})$, where the node features of $\mathbb{V}$ are $F^t_A$ (agent), $F^t_{l_k}$ (lane waypoint), $F^t_{v_k}$ (conventional vehicle), and $F_{\mathit{tl}_k}^t$ (traffic light), and $\mathbb{E}$ includes the edges between the agent node and other nodes. Each node contains the corresponding feature vectors of $z$ time steps, e.g., $F^t_{A}=[f_{A}^{t-z+1}, f_{A}^{t-z+2},\dots, f_{A}^t]$. Then, we encode these node features by the Gated Recurrent Unit (GRU)~\cite{chung2014empirical} as follows:
\begin{equation}
    \small
    \begin{gathered}
    h_A^\tau = \mathit{GRU}(f_A^\tau, h_A^{\tau-1}; W_1),\\
    h_{l_k}^\tau = \mathit{GRU}(f_{l_k}^\tau, h_{l_k}^{\tau-1}; W_2),\\
    h_{v_k}^\tau = \mathit{GRU}(f_{v_k}^\tau, h_{v_k}^{\tau-1}; W_3),\\
    h_{tl_k}^\tau = \mathit{GRU}(f_{tl_k}^\tau, h_{tl_k}^{\tau-1}; W_4),
\end{gathered}
\end{equation}
where $\tau \in \{t-z+1, t-z+2,\dots, t\}$, $W_1 \in \mathbb{R}^{5\times D_1}$, $W_2 \in \mathbb{R}^{4m\times D_1}$, $W_3 \in \mathbb{R}^{12\times D_1}$, and $W_4 \in \mathbb{R}^{4\times D_1}$ are the weight matrices. The hidden vectors of GRU default to zero vectors when $\tau$ is $t-z+1$. Finally, we take the hidden vectors (i.e., $h_{A}^t, h_{l_k}^t, h_{v_k}^t, h_{tl_k}^t$) at time step $t$ as the node embedding.



(2) Cross Attention.
After embedding each node in $G^t_k$, we utilize a cross attention machanism~\cite{chen2022vehicle} to aggregate $G^t_k$ into a feature vector $E^t_k$ based on its edge connection as follows:

\begin{equation}
    \small
E^t_k=\mathit{Softmax}(\frac{W_5h_A^t(W_6H^t_k)}{\sqrt{D_2}})(W_7H_k^t),
\end{equation}
where $H_k^t=[h_{l_k}^t, h_{v_k}^t, h_{tl_k}^t]$, $W_5 \in \mathbb{R}^{D_1\times D_2}$, $W_6 \in \mathbb{R}^{D_1\times D_2}$, and $W_7 \in \mathbb{R}^{D_1\times D_2}$ are the weight matrices for computing a query matrix, a key matrix, and a value matrix based on the attention mechanism~\cite{chen2022vehicle}, and $D_2$ is the embedding dimension of these matrices.

(3) Lane-wised Fusion.
We execute the embedding and aggregation on all star graphs, i.e., $G_{k-1}^t, G_k^t, G_{k+1}^t$. Afterwards, we can get three feature vectors $E_{k-1}^t$, $E_{k}^t$, and $E_{k+1}^t$, which have respectively aggregated the multi-modal features on the lane $l_{k-1}, l_k, l_{k+1}$. Finally, we use a multi-layer perceptron network to fuse the concatenation of these vectors into a state representation $E_s^t$ as follows:
\begin{equation}
    \small
    E_s^t = W_8(E^t_{k-1} \oplus E^t_k \oplus E^t_{k+1}),
\end{equation}
where $W_8 \in \mathbb{R}^{3D_2\times D_3}$ is the weight matrix.

\subsection{Parameterized Action Calculation}
\label{parameterized action}
In order to form an optimal action policy with large reward values, previous researchers propose the Q-learning method~\cite{mnih2015human} to construct a Q-table to store the expected reward value $Q$. The idea of this method is to query the Q-table for the action with the largest Q value according to the current state $s^t$. However, the query and update of the Q-table are time-consuming, and the dimension of the Q-table will increase dramatically if the number of state features and alternative actions increases. Therefore, considering the complexity and continuity of states and actions in our work, we use neural networks instead of the Q-table, which uses an actor network and a critic network to respectively compute actions and estimate Q values. Next, we will introduce the action structure and calculation of our framework.

\begin{figure}[!t]
    \centering
    \setlength{\abovecaptionskip}{0.1cm}
    \includegraphics[width=0.46\textwidth]{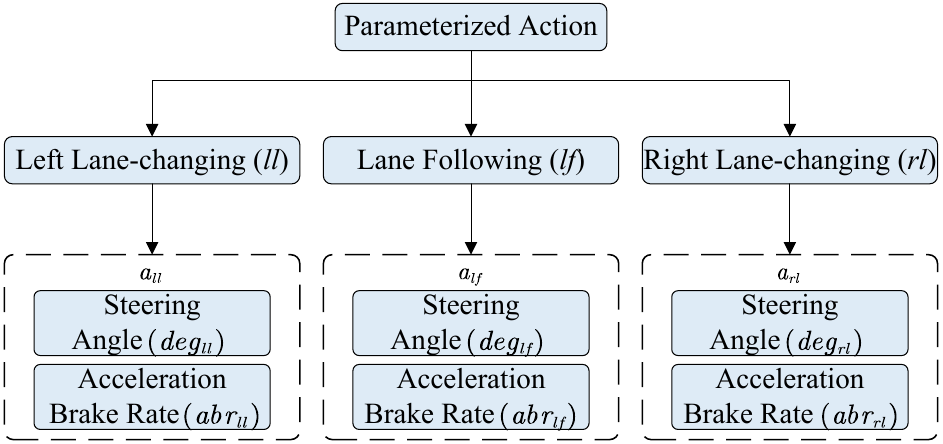}
    \caption{Parameterized Action}
    \label{action}
    \vspace{-0.6cm}
\end{figure}
 
\noindent \underline{\textbf{Parameterized Action.}}
Intuitively, an action of the agent in multi-lane scenarios requires a high-level decision (whether to perform lane-changing) and a low-level action (how to implement the high-level decision). A single steering angle is only suitable for open spaces with no lane division but not for multi-lane scenarios since it cannot distinguish between lane-changing behaviors and lane-following behaviors on curved roads~\cite{perez2021robot, xiao2022motion}.
Therefore, as shown in Figure~\ref{action}, we introduce a parameterized action structure to separate different lane-changing decisions. Specifically, we define three mutually exclusive high-level decisions $P$ comprising left lane-changing ($\small \mathit{ll}$), lane following ($\small \mathit{lf}$), and right lane-changing ($\small \mathit{rl}$), i.e., $\small P = (\mathit{ll}, \mathit{lf}, \mathit{rl})$. Each decision $p$ in $P$ is coupled with a low-level action $a_p$ that consists of a steering angle $\small \mathit{deg_p}$ and an acceleration brake rate $\small \mathit{abr_p}$, i.e., $\small a_p=(\mathit{deg_p}, \mathit{abr_p})$.

\noindent \underline{\textbf{Actor.}}
\label{actor_network}
Given the state $s^t$, the actor network is used to compute actions. For the network structure, it includes an LCA model in Section~\ref{LCA text} to learn a state representation $E^t_s$, based on which we use a multi-layer perceptron network (MLP) and an activation function $\mathit{Tanh}$ to compute an action vector $\small \mathcal{A}^t$ as follows:
\begin{equation}
\label{action calculation}
    \small
    \mathcal{A}^t=\mathit{Tanh}(W_9E_s^t), 
\end{equation}
where $W_9 \in \mathbb{R}^{D_3\times 6}$ is the weight matrix. $\mathcal{A}^t$ consists of all actions, i.e., $\small \mathcal{A}^t=[a_{\mathit{ll}}^t, a_{\mathit{lf}}^t, a_{\mathit{rl}}^t]$, where the steering angle and acceleration brake rate in each action will be scaled into the defined ranges in Section~\ref{definition}. 


\noindent \underline{\textbf{Q Value.}}
In this work, the goal of deep reinforcement learning is to select actions with maximal expected $\gamma$-discounted cumulative reward, i.e., the state-action value function Q~\cite{mnih2015human}. A better action is expected to have a larger Q value. When the agent performs an action $\small a^t_p$ of decision $p$ at the state $s^t$, the Q value $\small Q(s^t, p, a^t_p)$ is calculated using the Bellman Equation~\cite{watkins1992q}, as follows:
\begin{equation}
    \small
\begin{aligned}
&Q(s^t, p, a^t_p) := Q(s^t, \mathit{deg}_p^t, \mathit{abr}_p^t) = \mathbb{E}_{s^{t+1}}\big[r(s^t, a^t) + \\
&\gamma \max(\sup
Q(s^{t+1}, \hat{p}, a_{\hat{p}}^{t+1}) \big| a_{\hat{p}}^{t+1}=(\mathit{deg}_{\hat{p}}^{t+1}, \mathit{abr}_{\hat{p}}^{t+1}))\big],
\end{aligned}
\end{equation}
where $\gamma \in [0, 1)$ is a discount factor, the $\mathit{max}$ operation is used to find the best decision $\hat{p}$ from $P=(\mathit{ll},\mathit{lf},\mathit{rl})$, and the $\mathit{sup}$ is used to find the best action $a^{t+1}_{\hat{p}}$ under $\hat{p}$.

\noindent \underline{\textbf{Critic.}}
\label{critic_network}
Given the state $s^t$ and the action vector $\small \mathcal{A}^t$, we utilize a critic network to approximate a vector $\small Q(s^t, P, \mathcal{A}^t)$ that contains three Q values of the actions in $\mathcal{A}^t$, i.e., $ \small Q(s^t, P, \mathcal{A}^t)=[\mathit{Q(s^t, \mathit{ll}, a^t_{\mathit{ll}})}, \mathit{Q(s^t, \mathit{lf}, a^t_{\mathit{lf}})}, \mathit{Q(s^t, \mathit{rl}, a^t_{\mathit{rl}})}]$.
For the network structure, it first includes an LCA model in Section~\ref{LCA text} to learn a state representation $\small E_s^t$ and then uses a multi-layer perceptron network (MLP) and an activation function $\mathit{ReLU}$ to learn an action representation $\small E^t_a$ of $\small \mathcal{A}^t$ as follows:
\begin{equation}
\label{q calculation_1}
\small
    E_a^t=\mathit{ReLU}(W_{10}\mathcal{A}^t),
\end{equation}
where $W_{10} \in \mathbb{R}^{6\times D_4}$ is the weight matrix and $D_4$ is the dimension of $E_a^t$. 
Then, we apply another MLP to compute $\small Q(s^t, P, \mathcal{A}^t)$ with the concatenation of $\small E^t_s$ and $\small E^t_a$ as input, as follows:
\begin{equation}
\small
\label{q calculation_2}
    Q(s^t, P, \mathcal{A}^t) = W_{11}(E^t_s \oplus E^t_a),
\end{equation}
where $\small W_{11} \in \mathbb{R}^{(D_3+D_4)\times 3}$ is the weight matrix of the MLP.

\noindent \underline{\textbf{Action Selection.}}
After acquiring $\small \mathcal{A}^t$ and $\small Q(s^t, P, \mathcal{A}^t)$ from the actor network and critic network respectively, this component is utilized to select an action for the agent $A$. Based on a specific route of $A$, we first need to exclude actions that violate the route. It can be divided into two scenarios: (1) Mandatory lane-changing scenario. Taking the scenario (Figure~\ref{state generation}) mentioned in the state generation of Section~\ref{state generation text} as an example, the agent must change from the current lane to the left lane before the intersection. In this scenario, we will exclude the action of right lane-changing when the agent is less than 100 meters away from the endline of the current lane following ~\cite{shang2022off,huang2019determining}. The fake stationary vehicle on the endline will force the agent to perform left lane-changing in advance or stop and wait. (2) Non-mandatory lane-changing scenario. If the target lane of a lane-changing decision is not in the route, the action of this decision will be excluded. 

We exclude an action by setting its corresponding Q value to negative infinity. Then, we choose an action that has maximal Q value from all actions as follows:
\begin{equation}
    \begin{gathered}
    \label{selection}
    \hat{p} = \mathit{argmax}Q(s^t, P, \mathcal{A}^t),\\
    a^t := a^t_{\hat{p}}=(\mathit{deg}^t_{\hat{p}}, \mathit{abr}^t_{\hat{p}}),
\end{gathered}
\end{equation}
where $a^t_{\hat{p}}$ is the action that has maximal Q value. Therefore at time step $t$, the steering angle $\small \mathit{deg}^t_{\hat{p}}$ and acceleration brake rate $\small \mathit{abr}^t_{\hat{p}}$ is the action $a^t$ performed by the agent.

\begin{figure}[!t]
    \centering
    \setlength{\abovecaptionskip}{0.1cm}
    \includegraphics[width=0.48 \textwidth]{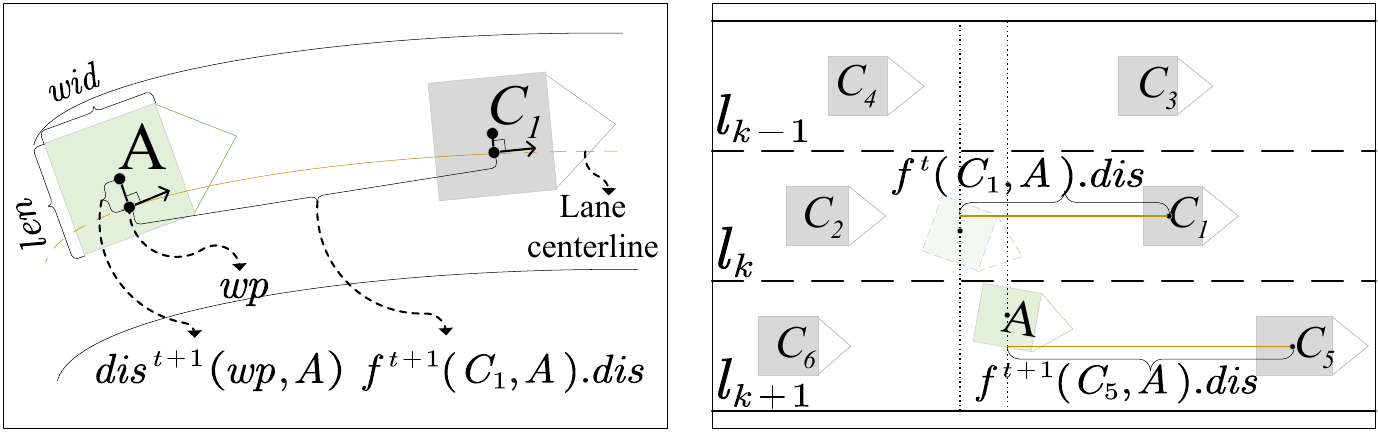}
    \caption{Example of Reward Calculation}
    \label{reward}
    \vspace{-0.6cm}
\end{figure}

\subsection{Hybrid Reward Function}
\label{reward function}
The reward function serves as an exploration signal to teach the agent to learn an optimal action policy. In this work, a good action of the agent $A$ should achieve safety, driving efficient, passenger comfort, and minimal impacts on other vehicles. Therefore, we construct a hybrid reward function considering four aspects as follows:

\begin{equation}
\label{reward formulation}
\small
r(s^t,a^t)=w_1r_1^t+w_2r_2^t+w_3r_3^t+w_4r_4^t, 
\end{equation}
where $w_1, w_2, w_3, w_4$ are four tunable coefficients to adjust the importance of safety, traffic efficiency, passenger comfort, and impact, respectively. Next, we define the reward values for safety ($r_1^t$), traffic efficiency ($r_2^t$), passenger comfort ($r_3^t$), and impact ($r_4^t$).

\noindent \underline{\textbf{Safety.}}
We set up three safety reward terms to penalize the agent $A$ for unsafe behaviors and traffic violations.

\noindent(\textbf{1}) Safety reward for vehicle.
This reward is set to avoid the agent from colliding with other vehicles and we use time-to-collision (TTC)~\cite{evans1991traffic,xia2022rise} as the safety indicator of the agent. As shown in the left of Figure~\ref{reward}, after $A$ performs an action $a^t$, $\small \mathit{TTC}^{t+1}$ is calculated as $\scriptsize \mathit{TTC}^{t+1}=\frac{f^{t+1}(C_1, A).\mathit{dis}-\mathit{len}_{\mathit{AC_1}}}{f^{t+1}(C_1, A).\mathit{vel}}$, where $\small f^{t+1}(C_1, A).\mathit{dis}$ denotes the relative distance between $C_1$ and $A$ along the lane, $\small \mathit{len}_{\mathit{AC_1}}$ equals to half the sum of the length of $A$ and $C_1$, and $\small f^{t+1}(C_1,A).\mathit{vel}$ denotes their relative velocity. Following~\cite{zhu2020safe}, if $\small \mathit{TTC}^{t+1}$ is between 0 and a threshold $\mathcal{G}$, the agent will receive a negative reward $r^t_{1,1}$ that is calculated as $\scriptsize r^t_{1,1}=\mathit{max}(-3, log(\frac{\mathit{TTC}^{t+1}}{\mathcal{G}}))$, otherwise $r^t_{1,1}=0$. 


\noindent(\textbf{2}) Safety reward for lane. This reward is set to avoid the agent from a large lane offset. As shown in the left of Figure~\ref{reward}, we first find the nearest waypoint $\mathit{wp}$ of the agent and then calculate the Euclidean distance $\small \mathit{dis^{t+1}(wp, A)}$ between $\mathit{wp}$ and $A$. Then, the safety reward $r_{1,2}^t$ for lane is calculated as $\small r^t_{1,2}=-\frac{
\mathit{dis^{t+1}(wp, A)}}{d}$, where $d$ is the lane width.

\noindent(\textbf{3}) Terminal conditions. We consider four terminal conditions for the agent: 1) collision with conventional vehicles, 2) out of the road, 3) route deviations, and 4) violation of traffic rules (i.e., traffic lights and signs). Once they happened, the current episode will be terminated and the agent receives a negative reward $r^t_{1,3}$ of -20.


\noindent \underline{\textbf{Traffic Efficiency.}}
We set up two traffic efficiency reward terms to reward high velocities and tactical lane-changing behaviors of the agent.

\noindent(\textbf{1}) Traffic Efficiency reward for velocity.
The velocity $\small f_A^{t+1}.\mathit{vel}$ of the agent directly reflects its driving efficiency. Thus, the efficiency reward value $r^t_{2,1}$ for velocity is defined as $\small r^t_{2,1}=\frac{f_A^{t+1}.\mathit{vel}}{\mathit{vel}_{\mathit{max}}}$, where $\small \mathit{vel}_{\mathit{max}}$ denotes the speed limit.

\noindent(\textbf{2}) Traffic Efficiency reward for tactical lane-changing.
A tactical lane-changing behavior refers to a maneuver where the agent attempts to avoid following a crowded lane~\cite{erdmann2015sumo}. As shown in the right of Figure~\ref{reward}, if the agent $A$ changes from a lane to an adjacent lane, we record the relative distance $\small f^t(C_1, A).\mathit{dis}$ between A and $C_1$ at time step $t$ and the relative distance $\small f^{t+1}(C_5, A).\mathit{dis}$ between $A$ and $C_5$ at time step $t+1$. Then, the efficiency reward $r^t_{2,2}$ for tactical lane-changing is as $\scriptsize r^t_{2,2}=(\frac{f^{t+1}(C_5, A).\mathit{dis}-\mathit{len}_{\mathit{AC_5}}}{f^t(C_1, A).\mathit{dis}-\mathit{len}_{\mathit{AC_1}}}-1)*20$. In this way, the agent will receive a positive reward if it changes from a crowded lane to a freer lane, otherwise a negative reward. 


To this end, we aggregate $r^t_{2,1}$ and $r^t_{2,2}$ to get the traffic efficiency reward $r^t_{2}$, i.e., $\small r^t_{2}=\sum^2_{i=1}r^t_{2,i}$.

\noindent \underline{\textbf{Passenger Comfort.}}
Jerk, defined as the change rate of acceleration, is used to measure comfort since it has a strong influence on the comfort of passengers~\cite{zhu2020safe}. 
The passenger comfort reward value $r^t_{3}$ is defined as $\small r^t_{3}=-\frac{|\mathit{acc}^t-\mathit{acc}^{t-1}|}{2\mathit{acc}_{\mathit{thr}}}$, where $\mathit{acc}^t$ and $\mathit{acc}^{t-1}$ are two accelerations of agent $A$ at time step $t$ and $t-1$, $\mathit{acc}_{\mathit{thr}}$ is an acceleration threshold.

\noindent \underline{\textbf{Impact.}}
The velocity change or lane-changing behavior of the agent could affect its rear vehicles, reducing their driving efficiency and comfort, especially for some harsh behaviors, e.g., hard braking and abrupt lane-changing~\cite{liu2021lane}. To reduce the impact of the agent, we design an impact reward value to measure the degree that the agent forces the rear vehicles to decelerate. Based on the analysis of existing studies~\cite{xia2022rise,liu2021lane}, whether a vehicle decelerates or changes lanes, it only affects the vehicle behind it. Thus, after the agent performs an action $a^t$, we record the deceleration of $C_2$ to calculate the impact reward $r_{4}^t$ as follows:

\begin{equation}
\small
r_{4}^t = \begin{cases}
\frac{f^{t+1}_{C_2}.\mathit{vel}-f^{t}_{C_2}.\mathit{vel}}{2\mathit{acc}_{\mathit{thr}} * \Delta t}, & f^{t}_{C_2}.\mathit{vel} - f^{t+1}_{C_2}.\mathit{vel} > \mathit{v}_{\mathit{thr}} \\
0, & \mathit{otherwise}, \\
\end{cases}
\end{equation}
where $C_2$ is the first vehicle behind the agent, $\mathit{v}_{\mathit{thr}}$ is a deceleration threshold used to measure whether the agent affects $C_2$, and $\small 2\mathit{acc}_{\mathit{thr}}*\Delta t$ refers to the possible maximum deceleration between two consecutive time steps. The impact reward $r_{4}^t$ only works when $C_2$ decelerates greater than $\mathit{v}_{\mathit{thr}}$ at time step $t+1$.

\subsection{Multi-worker Training}
\label{Multi-worker text}
In this section, we will introduce the training process of our reinforcement learning model and a multi-worker setting that is used to accelerate the training.

\noindent \underline{\textbf{Model Training.}}
In our decision-making task, the network model aims to optimize the parameterized action calculation in Section~\ref{parameterized action}, which consists of a discrete lane-changing decision and continuous action commands (i.e., steering angle and accelerate brake rate). 
Recently, a BP-DQN model~\cite{9835644} is proposed to directly learn the action policy in the parameterized action space, which has achieved the state-of-the-art performance. It concurrently outputs three actions corresponding to three different lane-changing decisions and then automatically selects the best action to execute, However, this method suffers from the overestimation problem of Q values, leading to suboptimal action policy and even training failure in our experiments. Therefore, we propose a RBP-DQN model that introduces a behavior cloning regularization technique~\cite{fujimoto2021minimalist} to solve the overestimation problem and thus improve the policy optimization. In RBP-DQN, the loss function of the actor network in Section~\ref{actor_network} is formulated as follows: 
\begin{equation}
\small
\label{train_actor}
        \mathcal{L}_a = -\sum_{p \in P}(\frac{\alpha Q(s^t, p, u_p(s^t; W_a))}{AvgQ}+(u_p(s^t; W_a)-a^t)^2),
\end{equation}
where $W_a$ denotes all the learnable parameters in the actor network, $u_p$ refers to the learned action policy, $\alpha$ is a tunable coefficient and $AvgQ$ is the average Q value in a batch of training experiences. Instead of just using the negative sum of Q values as the loss function, RBP-DQN first uses $\alpha$ to adjust the weight of the Q value $Q(s^t, p, u_p(s^t; W_a))$ and then add $(u_p(s^t; W_a)-a^t)^2$ into the loss function. It can reduce the shift between the learning policy and historical experiences and thus avoid the overestimation of Q values~\cite{fujimoto2021minimalist}.
Compared to other Q regularization methods~\cite{brandfonbrener2021offline,kostrikov2021offline}, this method fits the parameterized action structure perfectly and achieves stable optimization in our task.

In addition, the critic network remains the loss function in BP-DQN, which is formulated as follows:
\begin{equation}
\small
    \begin{gathered}
    \label{train_critic}
    \mathcal{L}_c = \frac{1}{2}(y-Q(s^t,p,a^t_p;W_c))^2, \\
\end{gathered}
\end{equation} 
where $W_c$ denotes all the learnable parameters in the critic network, $a^t_p$ is the action performed by the agent at state $s^t$, and $y$ is a target Q value calculated by a target actor network and a target critic network following~\cite{lillicrap2015continuous}.



\noindent \underline{\textbf{Multi-worker Setting.}}
In our experiments, we find that the environment interaction and update in the simulation take much longer time than the model update. Therefore, in order to make full use of computing resources and improve the training efficiency of our model, we employ multiple agents to interact with the traffic environment following the standard multi-worker setting in~\cite{bertsekas2021rollout,horgan2018distributed}. Specifically, there are $n$ workers and a learner in our framework. Each worker initializes an agent to asynchronously explore the environment with different random noises and maintains a separate replay buffer called \emph{LocalBuffer} to collect experiences. The learner is responsible for updating the network parameters (i.e., $W_c$ and $W_a$) and manages all experiences with a replay buffer called \emph{GlobalBuffer}. 


\begin{algorithm}
\small
    \caption{Worker}
    \label{algorithm1}
    \KwIn{The parameters $W_a$ of the actor network and $W_c$ of the critic network from the learner, the buffer size $B_l$ of \emph{LocalBuffer}}
    \Repeat{\text{convergence}}{
    $s^t \gets$ State generation (Section~\ref{state generation text})\;
    $\mathcal{A}^t \gets$ Actor($W_a$)\ (Equation~\ref{action calculation})\;
    $Q(s^t, P, \mathcal{A}^t)\gets$ Critic($W_c$)\ (Equation~\ref{q calculation_1} and~\ref{q calculation_2})\;
    $\hat{p}, a^t \gets$ Action selection\ (Equation~\ref{selection})\;
    $r(s^t, a^t)\gets$ Reward calculation (Equation~\ref{reward formulation})\;
    $s^{t+1} \gets$ Environment.step()\;
    \emph{LocalBuffer}.add(($s^t, \hat{p}, \mathcal{A}^t, r(s^t, a^t), s^{t+1}$))\;
    \If{LocalBuffer.\rm{size()}$\ \geq B_l$}{
$\mathit{exp} \gets$ \emph{LocalBuffer}.get()\;
    \emph{GlobalBuffer}.add($\mathit{exp}$)\;}
    }
\end{algorithm}
\vspace{-0.4cm}
\begin{algorithm}
\small
    \caption{Learner}
        \label{algorithm2}
    \KwIn{The size $b$ of mini-batches, the buffer size $B_g$ of \emph{GlobalBuffer}}
    \Repeat 
        {\text{convergence}}
    {
    $\mathit{EXP} \gets$ Sample $b$ experiences\;
    $\mathcal{L}_c$, $\mathcal{L}_a \gets$ Compute loss\ (Equation~\ref{train_actor} and~\ref{train_critic})\;
    $W_c$, $W_a \gets$ Update parameters\;
    }
\end{algorithm}

\section{Experiments\label{Experiments}}
\subsection{Experimental Settings}
We conduct our experiments on the CARLA simulator~\footnote{\url{https://carla.readthedocs.io/en/0.9.14/}}, which is a widely-used project focused on creating a publicly available virtual environment for the autonomous driving industry~\cite{Dosovitskiy17}. It has 8 available towns and supports almost all sensors with the goal of flexibility and realism in high-fidelity simulations. We use 7 towns for training and hold out Town05 for evaluation following the previous work TransFuser~\cite{prakash2021multi}. The training benchmark has 35 routes and the evaluation benchmark has 20 routes, each with various road structures, e.g., intersections. In each episode, the autonomous vehicle is required to follow the predefined routes without colliding, driving off the road, or violating red lights and speed limits. Further, we vary the traffic density in each episode from 60$\mathit{vehicles/km}$ to 180$\mathit{vehicles/km}$ and vary the weather condition (e.g., normal, rain, fog, and night) to demonstrate the scalability and robustness of our framework. Based on these environment variables, there are $16,800$ environment settings for training and $9,600$ environment settings for evaluation. 

\noindent \underline{\textbf{Implementation Details.}}
All of the experiments are run on an Ubuntu Server with an Intel(R) Xeon(R) Silver 4214 CPU @ 2.20GHz and 4 NVIDIA GeForce RTX 3080 GPUs. The detection range $D_{\mathit{tl}}$ of traffic lights by the camera is set to $60m$, and the detection range $D_{v}$ of conventional vehicles by LiDAR is set to $90m$.
The time interval $\Delta t$ between two decisions is set to $0.1s$~\cite{xia2022rise}.
For the network structure in the state representation and parameterized action calculation, we set the corresponding dimensions as $D_1=32$, $D_2=32$, $D_3=32$, and $D_4=32$. For the hybrid reward function, the TTC threshold $\mathcal{G}$ in the safety reward is set as 4$s$, and the acceleration threshold $\mathit{acc_{thr}}$ in the comfort reward is set to 3$m/s^2$ following the settings in the previous work~\cite{zhu2020safe}. In addition, the velocity change threshold $\mathit{v_{thr}}$ in the impact reward is set to $0.1m/s$, which is used to determine if driving behaviors of the autonomous vehicle affect its rear vehicle~\cite{qu2020jointly}. To the end, we train our model using the Adam optimizer~\cite{kingma2014adam} with a scheduled learning rate of 0.001 and a batch size of 128. The reward discount factor $\gamma$ in the Bellman function is 0.9 and the target actor network and critic network use a soft update mechanism with an updated ratio of 0.01 similar to DDPG~\cite{lillicrap2015continuous}. In the setting of multi-worker training, We employ 11 (i.e., $n=11$) workers to collect experiences and one learner to update parameters, which can make full use of the computing resource.

\noindent \underline{\textbf{Baselines.}}
We compare our framework with the following methods.

\noindent(\textbf{1})
ACC-LC~\cite{xiao2017realistic,erdmann2015sumo}. Rule-based method that includes an adaptive cruise algorithm and a traditional lane-changing model to make decisions.

\noindent(\textbf{2}) TransFuser~\cite{prakash2021multi}. 
End-to-End imitation learning-based method that takes the raw sensor data as input and uses a transformer-based network structure to imitate labeled human driving behaviors.

\noindent(\textbf{3}) DRL-MF~\cite{fu2022decision}. Deep reinforcement learning-based method that exploits useful features from image and point cloud data. It cannot perceive road maps and pose information (e.g., orientation) of the autonomous vehicle and ignore its impacts on the rear traffic flow.

\noindent(\textbf{4}) HEAD~\cite{9835644}. Deep reinforcement learning-based method that proposes a BP-DQN model to make decisions. It only computes discrete lane-changing decisions without continuous steering angle control. For a fair comparison, we integrate our state representation module into HEAD to learn the multi-modal features since it only utilizes vehicle features to make decisions.


\noindent \underline{\textbf{Variants.}}
In addition to these baselines, we also perform ablation experiments with some variants of our framework.

To evaluate the impact reward in the hybrid reward function in Section~\ref{reward function}, we set a variant to test its effectiveness in reducing the impact on the rear vehicle.  

\noindent(\textbf{1}) \myfw-NoIMP. We remove the impact reward value and only consider the safety, traffic efficiency, and passenger comfort reward values.

To evaluate our state representation module~\ref{LCA text}, we first set three variants to test its effectiveness and efficiency in exploiting useful features from multi-modal inputs.

\noindent(\textbf{2}) \myfw-NoVEC. Instead of vector representations of the multi-modal features in the state representation module, it rasterizes these features in a bird-eye view and uses convolutional layers to exploit useful features following~\cite{djuric2018short}.

\noindent(\textbf{3}) \myfw-NoSTAR. Instead of the agent-centric star graph in the state representation module, it constructs a complete graph to model the relationship between the agent and other traffic participants, i.e., lane waypoints, conventional vehicles, and traffic lights.

\noindent(\textbf{4}) \myfw-NoLANE. Instead of the lane-wised fusion paradigm in the state representation module, it directly fuses all feature vectors on different lanes by the cross attention layer.

Then, we set a variant to evaluate the robustness.

\noindent(\textbf{5}) \myfw-NoHIS. Instead of using historical features in the state generation module in Section~\ref{state generation text}, it only uses features at the current time step as the state. Afterwards, we use MLP to embed node features rather than GRU.

To evaluate the multi-worker training setting in Section~\ref{Multi-worker text}, we set a variant to test its effectiveness in improving the convergence performance of our framework.

\noindent(\textbf{6}) \myfw-NoMUL. Instead of multiple workers, it only uses one worker to explore the environment and store experiences.

\noindent \underline{\textbf{Compared Methods of Reinforcement Learning.}}
For our reinforcement learning model RBP-DQN, we compare it against three methods for optimizing our parameterized action calculation, as follows:

\noindent(\textbf{1}) Q-PAMDP~\cite{masson2016reinforcement}.
Parameterized Q-learning method that alternates learning discrete decisions and continuous action commands.

\noindent(\textbf{2}) P-DDPG~\cite{hausknecht2015deep}.
Parameterized deep deterministic policy gradients method that collapse the parameterized action into a continuous one.

\noindent(\textbf{3}) BP-DQN~\cite{9835644}.
Branched parameterized deep Q-network that directly learns the parameterized action structure without Q regularization.

\begin{table}[tp]
    \setlength{\abovecaptionskip}{0cm}
    \caption{Macroscopic Evaluation}
\small
    \centering

    \begin{tabular}{|p{2cm}<{\centering}|p{1.2cm}<{\centering}|p{1.2cm}<{\centering}|p{1.2cm}<{\centering}|}
         \hline
         \tabincell{c}{Methods}  &  \tabincell{c}{Avg \\ DT-A $(s)$} & \tabincell{c}{Avg \\ AT-C $(s)$} & \tabincell{c}{Avg \\ DI-C } \\
         \hline \hline
         ACC-LC & 57.08 & 7.03 & 1.46\\\hline
         TransFuser & 56.45 & 6.73 & 1.43\\\hline
         DRL-MF & 55.02 & 7.22 & 1.49 \\\hline
         HEAD & 55.57 & 6.55 & 1.41 \\\hline
         \myfw-NoIMP & \textbf{53.33} & 6.67 & 1.42 \\\hline
         \myfw & 53.61 & \textbf{5.48} & \textbf{1.35} \\\hline
         \end{tabular}

\vspace{-0.4cm}
\label{macro}
\end{table}

\subsection{Macroscopic Evaluation}
\label{Macroscopic}
In this section, we conduct a macroscopic evaluation of the autonomous vehicle regarding its traffic efficiency and impact on other vehicles. The compared methods consist of our framework \myfw, five baselines, and a variant \myfw-NoIMP.

\noindent \underline{\textbf{Evaluation Metrics}}
At each evaluation episode, all methods enable the autonomous vehicle to complete a route safely and without violation of traffic rules. Then, we utilize three metrics in each evaluation episode to evaluate the macroscopic effectiveness of our framework.

\noindent(\textbf{1})
Average driving time of the autonomous vehicle (AvgDT-A). We record the driving time of the autonomous vehicle for every 1$\mathit{km}$ road. A smaller AvgDT-A indicates that the autonomous vehicle has higher driving efficiency.

\noindent(\textbf{2})
Average affected time of the rear vehicle by the autonomous vehicle (AvgAT-C).
We record the total time that the rear vehicle decelerates greater than $0.1m/s$ within a time step (i.e., $0.1s$).
A smaller AvgAT-C indicates that the rear vehicle is less affected by the autonomous vehicle.

\noindent(\textbf{3})
Average delay index of the rear vehicle of the autonomous vehicle (AvgDI-C). Following ~\cite{xia2022rise}, the delay index (DI) refers to the ratio of the actual travel time to the ideal travel time in free state.

We report AvgDT-A, AvgAT-C, and AvgDI-C in Table~\ref{macro}. As depicted, our framework \myfw\ achieves a shorter AvgDT-A and AvgAT-C and has a smaller AvgDI-C compared to these baselines. These results demonstrate that \myfw\ not only enables the autonomous vehicle to have high driving efficiency but also reduces its impact on the rear vehicle. 
The first reason lies in that our framework \myfw\ can learn a comprehensive state representation that exploits useful features from multi-modal inputs.
Secondly, based on the learned representation, the parameterized action structure and hybrid reward function in \myfw\ enable the autonomous vehicle to learn tactical acceleration and lane-changing behaviors to achieve high traffic efficiency while minimizing the impact on other vehicles. 
To study the effectiveness of the impact reward term in reducing the impact caused by the autonomous vehicle, we compare our framework \myfw\ against a variant \myfw-NoIMP that removes the impact reward term. As shown in Table~\ref{macro}, compared to \myfw, \myfw-NoIMP has a slightly better AvgDT-A but has a noticeably worse AvgAT-C and AvgDI-C. This is because the autonomous vehicle under \myfw-NoIMP performs harsh actions in pursuit of high traffic efficiency but is prone to trigger a large impact and delay to the rear vehicle.
Therefore, the impact reward term is critical in lowering the impact caused by our autonomous vehicle.

\begin{table}[tp]
    \setlength{\abovecaptionskip}{0cm}
\caption{Microscopic Evaluation}
\small
    \centering
    \begin{tabular}{|p{2cm}<{\centering}|p{0.84cm}<{\centering}|p{0.84cm}<{\centering}|p{0.84cm}<{\centering}|p{0.84cm}<{\centering}|p{0.84cm}<{\centering}|}
         \hline
         \tabincell{c}{Methods} & \tabincell{c}{Min\\TTC-A \\ $(s)$} & \tabincell{c}{Avg\\OFF-A \\ $(m)$} & \tabincell{c}{Avg\\VEL-A \\ $(m/s)$} & \tabincell{c}{Avg\\JERK-A \\ $(\mathit{m/s^2})$} & \tabincell{c}{Avg\\DEC-C \\ $(m/s)$} \\
         \hline \hline
         ACC-LC & 2.94 & 0.19 & 17.99 & 1.32 & 0.55\\\hline
         TransFuser & 2.55 & 0.31 & 18.35 & 1.17 & 0.52\\\hline
         DRL-MF & 2.78 & 0.37 & 18.81 & 1.40 & 0.59\\\hline
         HEAD & 2.85 & 0.20 & 18.48 & 1.05 & 0.48\\\hline
         \myfw-NoIMP & 2.23 & 0.22 & \textbf{19.25} & 1.07 & 0.50\\\hline
         \myfw & \textbf{3.15} & \textbf{0.17} & 19.12 & \textbf{0.97} & \textbf{0.44}\\\hline
         \end{tabular}
\label{micro}
\vspace{-0.2cm}
\end{table}

\subsection{Microscopic Evaluation}
\label{Microscopic}
In this section, we conduct a microscopic evaluation of our framework \myfw, five baselines, and a variant \myfw-NoIMP for each action of the autonomous vehicle.

\noindent \underline{\textbf{Evaluation Metrics}}
We utilize five metrics to evaluate the microscopic effectiveness based on our hybrid reward function (Section~\ref{reward function}).

\noindent(\textbf{1})
Minimal time-to-collision of the autonomous vehicle (MinTTC-A). We record the time-to-collision (TTC) of the autonomous vehicle. A larger MinTTC-A indicates that
the autonomous vehicle is safer.

\noindent(\textbf{2})
Average lane offset of the autonomous vehicle (AvgOFF-A). We record the offset related to the lane centerline when the autonomous vehicle follows a lane. A smaller AvgOFF-A indicates that the vehicle performs better in lane following.

\noindent(\textbf{3})
Average velocity of the autonomous vehicle (AvgVEL-A).
We record the velocity of the autonomous vehicle. A larger AvgVEL-A indicates higher traffic efficiency.

\noindent(\textbf{4})
Average jerk of the autonomous vehicle (AvgJERK-A).
We record the acceleration change of the autonomous vehicle. A smaller AvgJERK-A indicates a more comfortable driving experience.

\noindent(\textbf{5})
Average deceleration of the rear conventional vehicle of the autonomous vehicle (AvgDEC-C). After the autonomous vehicle performs braking or lane-changing behaviors, we record the deceleration of the vehicle behind it. 
A smaller AvgDEC-C means the autonomous vehicle has less impact on the vehicle behind it.

We report MinTTC-A, AvgOFF-A, AvgVEL-A, AvgJERK-A, and AvgDEC-C in Table~\ref{micro}. Compared to four baselines, we can see that our framework \myfw\  has the longest MinTTC-A, the highest AvgVEL-A, and the smallest AvgOFF-A, AvgJERK-A, and AvgDEC-C, demonstrating that \myfw\ enables the autonomous vehicle to generate acceleration brake rates and steering angles with safety, traffic efficiency, passenger comfort, and minimal impact on the rear vehicle. In these baselines, both DRL-MF and HEAD outperform the rule-based method ACC-LC and the imitation learning-based methods TransFuser due to the superior optimization effect of reinforcement learning.
For DRL-MF, it lacks consideration of the road semantic information and pose information (e.g., position and orientation) of the autonomous vehicle and thus cannot accurately estimate the relative pose information between the autonomous vehicle and the specific route, leading to a poorer performance of DRL-MF compared to \myfw\ in all metrics, especially the lane offset. In addition, the experimental results of HEAD and \myfw\ prove that our parameterized action structure and reinforcement learning model RBP-DQN facilitate the decision-making of the autonomous vehicle.
Further, the experimental results of \myfw-NoIMP prove that removing the impact reward item will lower all metrics except traffic efficiency from a microscopic point of view.

\begin{table}[h]
\setlength{\abovecaptionskip}{0cm}
\caption{Effect of State Representation}
\renewcommand\arraystretch{1.15}
\small
\centering
\begin{tabular}{|p{1.5cm}<{\centering}|p{1.1cm}<{\centering}|p{1.1cm}<{\centering}|p{1.1cm}<{\centering}|p{1.1cm}<{\centering}|}
\hline
 Metric & \tabincell{c}{\myfw-\\NoVEC} & \tabincell{c}{\myfw-\\NoSTAR} & \tabincell{c}{\myfw-\\NoLANE} & \myfw \\
\hline
AvgR & 0.31&0.46&0.45&\textbf{0.48}\\ \hline
 AvgIT$\ (ms)$ & 51.83 & 17.08 & \textbf{8.62} & 8.96 \\
\hline
\end{tabular}
\vspace{-0.3cm}
\label{infer}
\end{table}

\subsection{Evaluation of State Representation}
\label{Representation}
The lane-wised cross attention model (LCA) in our state representation module is proposed to exploit valuable features from multi-modal inputs. 
We calculate the average reward value (AvgR) of each action in the evaluation phase to compare the effectiveness of \myfw\ against three variants \myfw-NoVEC, \myfw-NoSTAR, and \myfw-NoLANE in Table~\ref{infer}. As depicted, \myfw-NoVEC has the smallest AvgR among these methods, which proves that vector-based methods are more effective than raster-based methods in representing the multi-modal environment. Then, We can see that \myfw\ has a higher AvgR compared to both \myfw-NoSTAR and \myfw-NoLANE, which demonstrates that both the agent-centric star graph structure and lane-wised fusion paradigm in \myfw\ are beneficial in learning useful semantic features from the multi-modal traffic and thus improving the decision-making performance of the autonomous vehicle.  

\noindent \underline{\textbf{Inference time.}}
In our decision-making framework, the average inference time (AvgIT) to calculate each action depends mainly on the time required to compute a representation of observed state features. We present the AvgIT in table~\ref{infer}. As shown, \myfw\ has a higher decision-making speed than \myfw-NoVEC and \myfw-NoSTAR and a close speed to \myfw-NoLANE. Further, compared to the time interval ($\Delta t=0.1s=100ms$) between two decisions, the inference time of $\small 8.96ms$ can meet the requirement of real-time decision-making.




\begin{figure}[!t]
\centering
\begin{minipage}[t]{0.23\textwidth}
\centering
\setlength{\abovecaptionskip}{0.1cm}
\includegraphics[width=4.2cm]{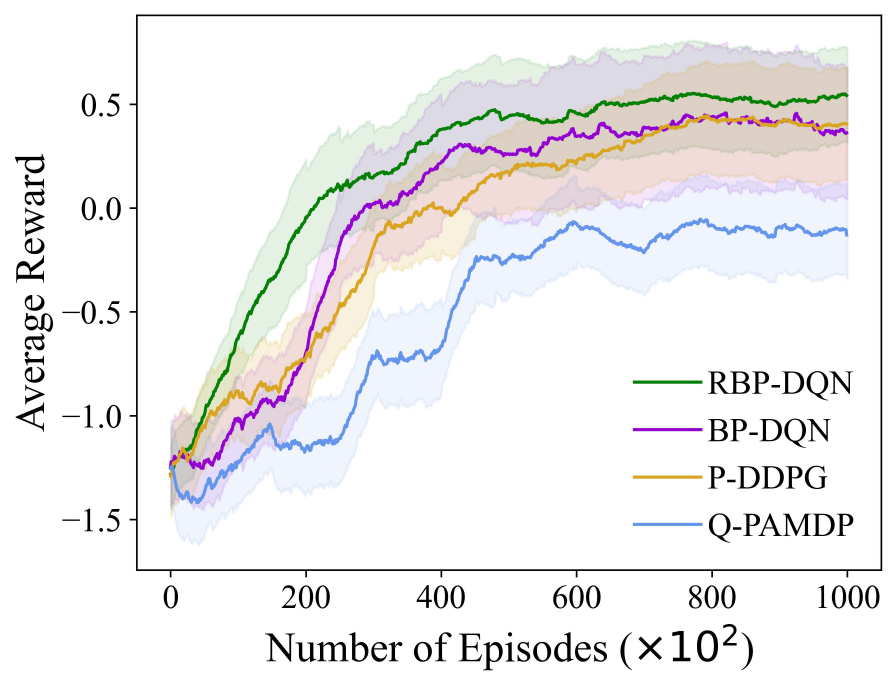}
\caption{Effect of Reinforcement Learning}  
\label{state evaluation}
\end{minipage}
\begin{minipage}[t]{0.23\textwidth}
\centering
\setlength{\abovecaptionskip}{0.1cm}
\includegraphics[width=4.2cm]{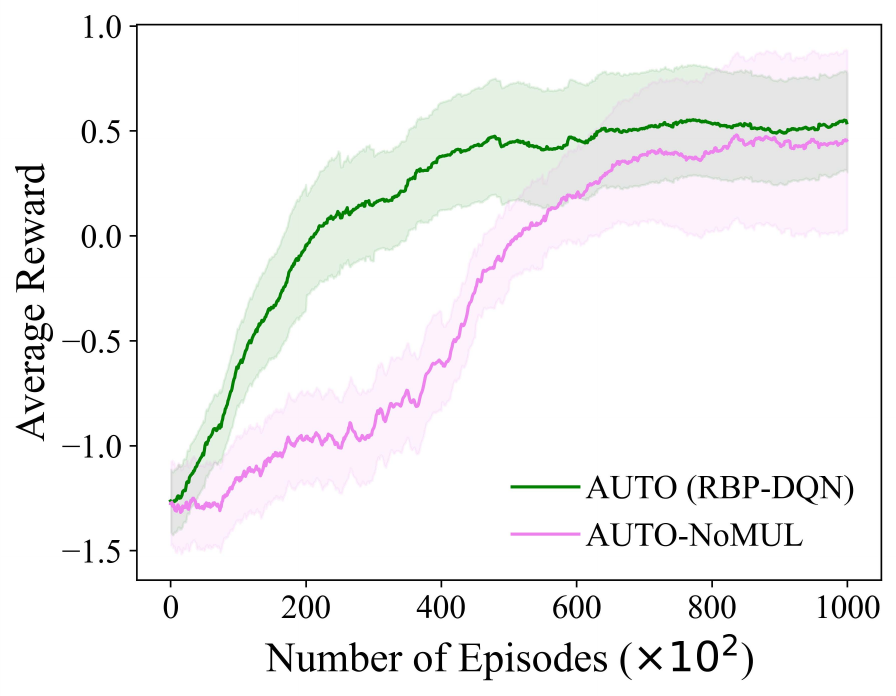}
\caption{Effect of Multi-worker Setting}
\label{multi-worker}
\end{minipage}
\vspace{-0.4cm}
\end{figure}
nk

\begin{figure*}[thbp]
\centering
\subfloat[ Driving Time.]{
\begin{minipage}[t]{0.23\linewidth}
\centering
\includegraphics[width=0.98\textwidth]{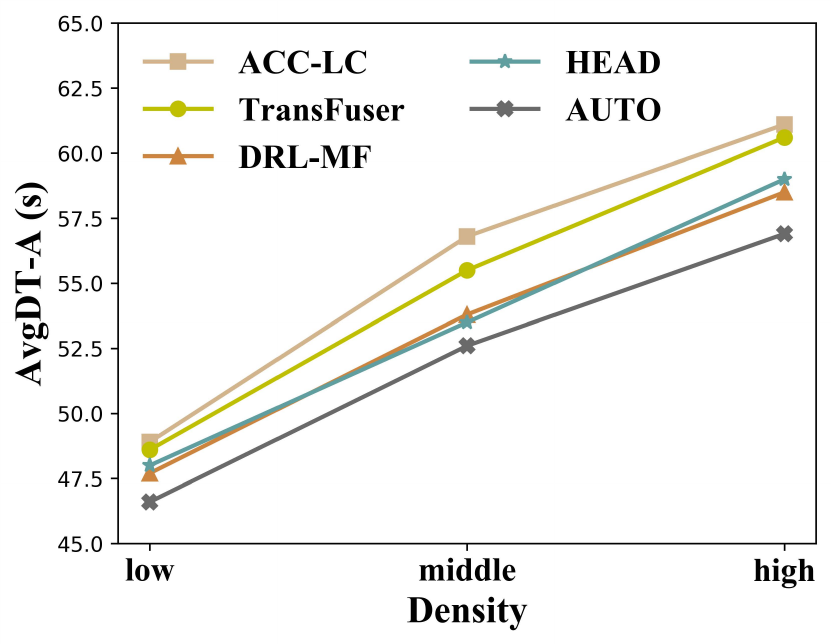}
\end{minipage}%
}%
\subfloat[ Affected Time.]{
\begin{minipage}[t]{0.23\linewidth}
\centering
\includegraphics[width=0.94\textwidth]{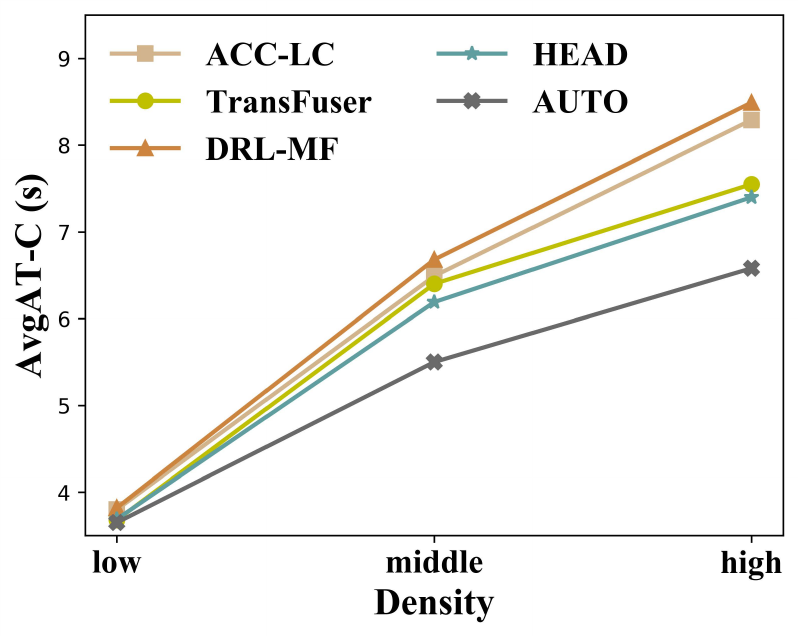}
\end{minipage}%
}%
\subfloat[ Delay Index.]{
\begin{minipage}[t]{0.23\linewidth}
\centering
\includegraphics[width=0.98\textwidth]{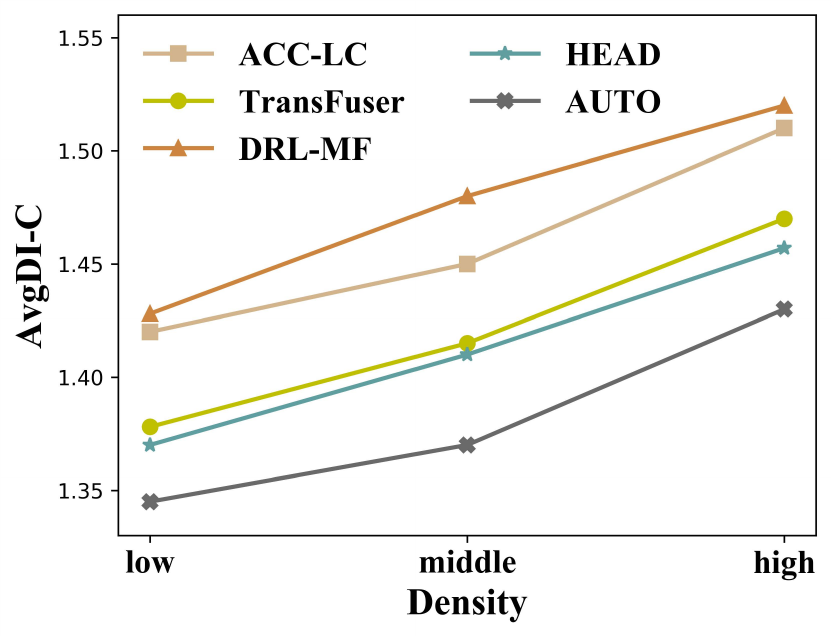}
\end{minipage}
}%
\subfloat[Driving Time of Traffic Flow.]{
\begin{minipage}[t]{0.23\linewidth}
\centering
\includegraphics[width=0.97\textwidth]{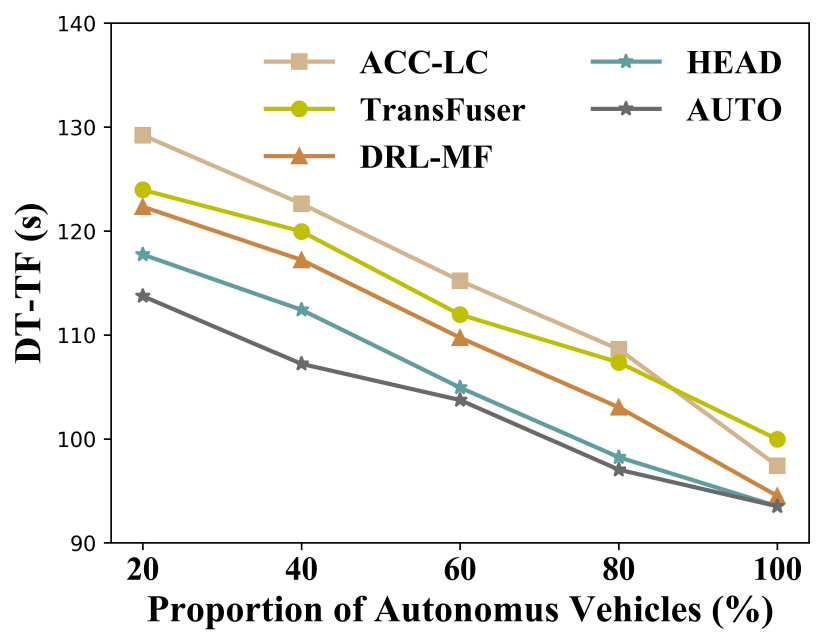}
\end{minipage}
}%
\centering
\caption{Scalability Analysis}
\label{Scalability Analysis}
\end{figure*}

\subsection{Evaluation of Reinforcement Learning}
\label{reinforcement evaluation}
To evaluate the proposed reinforcement learning model RBP-DQN in our framework, we compare it with Q-PAMDP, P-DDPG, and BP-DQN. As shown in Figure~\ref{state evaluation}, we provide the learning curves of these methods. We can see that RBP-DQN has the highest convergence reward and converges faster than other methods. The reason for its superior optimization performance is that it can effectively learn the parameterized action structure and can avoid the overestimation problem of Q values via a regularization technique. For the regularization term, we find that our framework \myfw\ reaches the peak performance when $\alpha$ equals 0.6. Unless otherwise specified, other experiments are done when $\alpha$ equals 0.6.




\subsection{Evaluation of Multi-worker Training}
The multi-worker setting in our framework \myfw\ is utilized to speed up the training process and increase the diversity of learnable experiences. As shown in Figure~\ref{multi-worker}, to evaluate the effectiveness, we compare the learning curves of \myfw\ against the variant \myfw-NoMUL that removes the multi-worker setting. We can see both the convergence speed and convergence reward of \myfw\  are better than \myfw-NoMUL. This is because the joint exploration by multiple agents can collect diverse experiences and explore different regions of the solution space, thus improving the learning efficiency of \myfw\ and avoiding getting stuck in local optima.

\begin{table}[t]
\setlength{\abovecaptionskip}{0cm}
\caption{Robustness under Different Weather Conditions}
\renewcommand\arraystretch{1.15}
\footnotesize
\centering
\begin{tabular}{|c|c|c|c|c|}
\hline
PT-RT (\%) & \multicolumn{4}{c|}{Weather Conditions} \\
\hline
Methods & Normal & Rain & Fog & Night \\
\hline
\hline
\myfw-NoHIS & 2.2 & 9.8 & 12.1 & 2.9 \\
\hline
\myfw & 1.8 & 6.3 & 10.3 &  2.7 \\
\hline
\end{tabular}
\label{Robustness table}
\end{table}

\subsection{Robustness Analysis}
To evaluate the robustness of our framework \myfw\ to observation noises, we provide the safety performance of the autonomous vehicle in four weather conditions, i.e., normal, rain, fog, and night. Under these conditions, the observed features undergo a 3\% to 18\% perturbation compared to the ground truth. For example, the traffic light detection in our experiment has an average error rate of 13\% in rainy conditions. To evaluate the robustness, we record the proportion of time (PT-RT) that the autonomous vehicle is under a risky time-to-collision (TTC). Based on the previous study~\cite{xia2022rise}, a vehicle can be considered to be at risk of collision when its TTC value is less than $4s$. As shown in Table~\ref{Robustness table}, we report the PT-RT of our framework \myfw\ and a variant \myfw-NoHIS that only takes the current environmental features as states but disregards the historical features. As shown, \myfw\ always has a smaller PT-RT than \myfw-NoHIS under all challenging weather conditions, which proves that incorporating the historical environmental features into states is beneficial to the robustness of our framework.


\subsection{Scalability Analysis}
In this work, we train our framework \myfw\ on a training benchmark and evaluate it on an evaluation benchmark. From the macroscopic evaluation in Section~\ref{Macroscopic} and the microscopic evaluation in Section~\ref{Microscopic}, we can see that \myfw\ can achieve a good performance in the evaluation scenarios, demonstrating a great potential of \myfw\ in terms of scaling to unseen scenarios. Furthermore, we conduct two experiments to evaluate our framework when the density of conventional vehicles changes or the proportion of autonomous vehicles increases.

\noindent \underline{\textbf{Increase Conventional Vehicles.}}
In Figure~\ref{Scalability Analysis}(a), (b), and (c), we use three macroscopic metrics (i.e., average driving time AvgDT-A, average affected time AvgAT-C, and average delay index AvgDI-C that are defined in Section~\ref{Macroscopic}) to evaluate the driving performance of the autonomous vehicle under three different densities (low: 60 vehicles/km, middle: 120 vehicles/km, and high: 180 vehicles/km) of conventional vehicles. Evidently, as the density increases, the autonomous vehicle has a decreasing traffic efficiency and is more prone to trigger large negative impacts on the traffic flow. As shown, based on the comparison of our framework \myfw\ with the baselines, we can see that \myfw\ always performs better, which proves its superiority in adaptation to different traffic densities.

\noindent \underline{\textbf{Increase Autonomous Vehicles.}}
In addition to the driving performance of an autonomous vehicle, we further explore the traffic efficiency of an entire traffic flow when the proportion of autonomous vehicles increases. Specifically, as the proportion scales from 20\% to 100\%, we record the driving time (DT-TF) of the entire traffic flow of $\mathit{1km}$. As shown in Figure~\ref{Scalability Analysis}(d), our framework \myfw\ always has a smaller DT-TF compared to the baselines. Moreover, as the proportion increases, the DT-TF of \myfw\ progressively reduces, demonstrating that the increasing proportion of autonomous vehicles is beneficial to traffic flow efficiency.

\begin{table}[h]
\setlength{\abovecaptionskip}{0cm}
\caption{Effect of Coefficients in Hybrid Reward Function}
\renewcommand\arraystretch{1.15}
\footnotesize
\centering
\begin{tabular}{|c|c|c|c|c|}
\hline
Coefficient & Min & Max & Step & Best \\
\hline
\hline
$w_1$ & 0.5 & 1 & 0.1 &  \textbf{0.9} \\
\hline
$w_2$ & 0 & 1 & 0.2 &  \textbf{0.8} \\
\hline
$w_3$ & 0 & 1 & 0.2 &  \textbf{0.6} \\
\hline
$w_4$ & 0 & 0.5 & 0.1 &  \textbf{0.2} \\
\hline
\end{tabular}
\label{reward shaping}
\end{table}

\subsection{Reward Shaping}
\label{reward shaping}
For the hybrid reward function in Section~\ref{reward function}, $w_1$, $w_2$, $w_3$, and $w_4$ correspond to the coefficients of $r_1$ (safety), $r_2$ (traffic efficiency), $r_3$ (passenger comfort), and $r_4$ (impact),
respectively. To achieve better performance, we adopt the grid search strategy~\cite{syarif2016svm} on a small number of training scenarios to determine them, which is shown in Table~\ref{reward shaping}. Unless otherwise specified, we set the coefficients as $w_1 = 0.9$, $w_2 = 0.8$, $w_3 = 0.6$, $w_4 = 0.2$ in all experiments.

\subsection{Hyper-Parameter Sensitivity}
\label{Hyper-Parameter Sensitivity}
In this section, we study the effect of different detection distances of LiDAR and Camera since they can impact the decision-making performance of the autonomous vehicle. Intuitively, Further detection distances can benefit autonomous vehicle decision-making. However, in real autonomous vehicle systems, further detection results are always accompanied by greater detection noise. Moreover, distant detection results usually have little impact on decision-making. As shown in Table~\ref{lidar range} and ~\ref{camera range}, we use the average reward of all actions to represent model performance. The best detection distances of LiDAR and Camera in our experiments are 90 meters and 60 meters respectively.

\begin{table}[h]
\footnotesize
\setlength{\abovecaptionskip}{0cm}
\caption{Impact of Detection Range of LiDAR}
\renewcommand\arraystretch{1.15}
\centering
\begin{tabular}{|p{2cm}<{\centering}|p{0.5cm}<{\centering}|p{0.5cm}<{\centering}|p{0.5cm}<{\centering}|p{0.5cm}<{\centering}|p{0.5cm}<{\centering}|p{0.5cm}<{\centering}|p{0.5cm}<{\centering}|}
\hline
 Detection Range & \tabincell{c}{50m} & \tabincell{c}{60m} & \tabincell{c}{70m} & 80m & 90m & 100m &110m\\
\hline
Average Reward & 0.454&0.472&0.481&0.482&0.483&0.482&0.482\\
\hline
\end{tabular}
\label{lidar range}
\end{table}

\begin{table}[h]
\footnotesize
\setlength{\abovecaptionskip}{0cm}
\caption{Impact of Detection Range of Camera}
\renewcommand\arraystretch{1.15}
\centering
\begin{tabular}{|p{2cm}<{\centering}|p{0.5cm}<{\centering}|p{0.5cm}<{\centering}|p{0.5cm}<{\centering}|p{0.5cm}<{\centering}|p{0.5cm}<{\centering}|p{0.5cm}<{\centering}|p{0.5cm}<{\centering}|}
\hline
 Detection Range & \tabincell{c}{50m} & \tabincell{c}{60m} & \tabincell{c}{70m} & 80m & 90m & 100m &110m\\
\hline
Average Reward & 0.481&0.483&0.481&0.481&0.480&0.480&0.479\\
\hline
\end{tabular}
\label{camera range}
\end{table}

\section{Related Work}
\label{related work}
Autonomous driving is considered to be one of those technologies that could herald a major shift in transportation~\cite{schrank2021urban}. 
In the following, we will introduce some related works on feature extraction and decision-making in autonomous driving.

\noindent
\underline{\textbf{Feature Extraction.}}
In this paper, we have built a fully-autonomous driving perception stack that can obtain the real-time multi-modal features of lanes, vehicles, and traffic lights from an HD map and onboard sensors. We conduct the experiments in a widely-used simulator Carla that mimics real-world traffic scenarios. It provides high-fidelity sensor models that can replicate the characteristics (e.g., field of view and resolution) of real sensors and generate synthetic data based on the virtual environment and the positions of objects. Specifically, we use an open-source standard OpenDrive~\cite{diaz2022hd} as the format of HD maps and estimate the pose of autonomous vehicles by fusing the geographic position provided by GNSS and the orientation provided by IMU~\cite{ribeiro2004kalman}. Then, we extract the features of surrounding vehicles from the point cloud data by LiDAR and extract the features of traffic lights from the image data by Camera.

For real-time analysis, with the development of feature extraction technologies, researchers have developed lightweight models that maintain high accuracy but require less computational power. For example, the methods in~\cite{wang2023yolov7,lari2011adaptive} can obtain the position and velocity information of surrounding vehicles at greater than 30 fps. In order to improve the robustness, our framework incorporates the detection results at multiple time steps to alleviate the impacts of detection missing and errors.


\noindent
\underline{\textbf{Decision-making.}}
Automatic driving decision-making methods can be categorized into end-to-end~\cite{prakash2021multi,jia2023thinktwice, coelho2022review, wu2022trajectory}, and modular approaches~\cite{leurent2019social,aradi2020survey,nageshrao2019autonomous,fu2022decision,zhu2020safe,xu2021patrol} based on their underlying architecture and design principles. End-to-end methods directly map raw sensor data to control commands, typically using deep neural networks. For example, TransFuser~\cite{prakash2021multi} proposes a transformer-based network structure to imitate labeled human driving behaviors. To improve safety, ThinkTwice~\cite{jia2023thinktwice} injects spatial-temporal prior knowledge and dense supervision into the imitation learning process. In contrast, modular methods firstly extract vehicle and traffic light features from raw sensor data and then calculate control commands. Compared with training directly based on raw sensor data, modular methods have better interpretability and decision-making performance since they start training from refined detection results rather than redundant sensor data.
In this paper, we focus on the modular methods. However, existing modular methods either ignore the complexity of environments only fitting straight roads, or ignore the impact on surrounding vehicles during the optimization phase. To address these limitations, we aim to design a decision-making framework that can not only achieve safety, traffic efficiency, passenger comfort, and minimal impact on traffic flow, but also adapt to complex traffic scenarios with multi-modal features.

Although the above learning-based methods can perform well in fusing multi-modal features and multi-objective optimization, there is always a safety module to guarantee safety in industrial autonomous driving systems~\cite{wang2019lane, krasowski2020safe,muhammad2020deep}. Specifically, the safety module is used to check whether the action output by the learning-based decision-making module is safe. If it is unsafe, the autonomous vehicle enters safe mode and uses a rule-based method to calculate a driving action. This usually occurs in emergency situations, such as a person suddenly appearing in front of the vehicle.

\section{conclusion\label{conclusion}}
In this paper, we propose \myfw\ to enable the autonomous vehicle to complete a specific route with safety, traffic efficiency, passenger comfort, and minimal impact on the rear vehicle. We first conduct a comprehensive perception of the traffic environment to obtain multi-modal inputs for the autonomous vehicle. Then, we design a graph-based model to exploit useful semantic features from the input and use a parameterized action paradigm to calculate fine-grained actions based on coarse-grained decisions. Finally, we propose a hybrid reward function to guide the optimization of reinforcement learning and use a regularization term
and a multi-worker setting to enhance the training.
Extensive experiments confirm the superiority of \myfw\  over state-of-the-art approaches in terms of both macroscopic and microscopic effectiveness.


\section*{Acknowledgment}
This work is partially supported by NSFC (No. 62272086, 61972069, 61836007 and 61832017),  Shenzhen Municipal Science and Technology R\&D Funding Basic Research Program (JCYJ20210324133607021), and Municipal Government of Quzhou under Grant (No. 2022D037, 2023D004, 2023D044), and Key Laboratory of Data Intelligence and Cognitive Computing, Longhua District, Shenzhen.

\bibliographystyle{IEEEtran}
\bibliography{IEEEabrv,sample}

\end{document}